\documentclass[letterpaper]{article} 
\usepackage[preprint]{aaai2027}  
\usepackage[hyphens]{url}  
\usepackage{graphicx} 
\usepackage{natbib}  
\usepackage{caption} 
\usepackage{algorithm}
\usepackage{algorithmic}
\usepackage{amsmath}
\usepackage{amssymb}
\usepackage{enumitem}
\usepackage{subcaption}
\usepackage{multirow, makecell}
\usepackage{color} 
\usepackage{soul}  

\usepackage{newfloat}
\usepackage{listings}
\DeclareCaptionStyle{ruled}{labelfont=normalfont,labelsep=colon,strut=off} 
\floatstyle{ruled}
\newfloat{listing}{tb}{lst}{}
\floatname{listing}{Listing}

\usepackage{booktabs}

\title{Tokenizer-Agnostic Engram Module}
\author{
    Jia Peng Lim{\rm 1}
}
\affiliations{
    \textsuperscript{\rm 1}Singapore Management University\\

    1101 Pennsylvania Ave, NW Suite 300\\
    Washington, DC 20004 USA\\
    proceedings-questions@aaai.org
}

\author {
    Jia Peng Lim\textsuperscript{\rm 1,\rm 2}
    Hai Leong Chieu\textsuperscript{\rm 2}
}
\affiliations {
    \textsuperscript{\rm 1}Singapore Management University\\
    \textsuperscript{\rm 2}DSO National Laboratories\\
    jiapeng.lim.2021@smu.edu.sg, chaileon@dso.org.sg
}

\begin{document}

\maketitle

\begin{abstract}
Deepseek's Engram, a conditional memory module, was introduced to trade-off storage versus reasoning in large language models. However, the module relies on token-level $N$-gram hashing for Engram embedding lookup, introducing a tight coupling to the tokenizer used:
a model with a different tokenizer would have to train its own Engram embeddings from scratch.
To improve the reusability of Engram embeddings, we propose a change to the hashing routine, enabling compatibility between Engram models using different tokenizers. 
Instead of modelling disjoint $N$-gram spaces, we treat $N$-gram as a method to sample potentially useful byte sequences, from all possible byte sequences across tokens.
We replace the XOR-based hashing with the general polynomial hashing with a joint embedding space across $N$. 
This work investigates the possible trade-offs and shows that this simple substitution produces comparable performance and achieves tokenizer-agnosticism: hash equivalence for byte-equivalent token sequences.
\end{abstract}

\begin{links}
    \link{Code}{https://github.com/jararap/polyhash-engram}
\end{links}

\section{Introduction}
\paragraph{DeepSeek's Engram Module.} 
Their influential work \cite{cheng2026conditionalmemoryscalablelookup} proposes using token-level $N$-gram conditional embedding lookup as an efficient parameter store. 
Each token has an additional vector input, depending on the previous $N$-1 token sequence\footnote{Pad tokens are placeholders for the tokens at the start.}, with each value acting as a hash key to the Engram embedding table. 
For each $n \in (1,N]$, multiple embedding vectors (heads) are sampled from the corresponding embedding table, i.e., each $n$-gram space is disjoint. 
These concatenated Engram embeddings along $N$ serve as the key/value input to the scaled dot product attention (SPDA) \cite{vaswani2017attentionneed}, with the query being the residual of the previous model block.
The context-aware SPDA gates the retrieved embeddings, before the 1D convolution operation along the token-level axis and adding the result back to the residual stream.

\paragraph{Tokenizer lock in.} Tokenizers may share the same algorithm, e.g., Byte-Pair Encoding (BPE) \cite{gage-1994-new, sennrich-etal-2016-neural} or Unigram \cite{kudo-richardson-2018-sentencepiece, kudo-2018-subword}, but may differ in their vocabulary set and size, resulting in different token sequence inputs.
Each token has a corresponding initial hash value with the final hash key(s) being an aggregation of these token hashes in a rolling XOR-wise manner. 
This causes a tight coupling between the Engram embeddings to the tokenizer used.
To share Engram embeddings across different models, they must use the same tokenizer. 
Model performance scales with tokenizer vocabulary size \cite{tao2024scaling, takase2025large}, so being able to adjust the tokenizer, with respect to model size or architecture, is an important consideration.
For models with fewer parameters, a fixed tokenizer requirement hurts the flexibility to control the ratio of embedding to feedforward parameters.

\paragraph{$N$-gram modelling an illusion?} $N$-grams are widely studied and used to evaluate and model local information \cite{brants2007large, buck2014n, liu2024infinigram, nguyen2024understanding, merrill2024evaluating}. 
We argue that what is crucial is the underlying byte sequence of the $N$-grams rather than modelling specific $n$-gram tokens.
Consider this case where our two tokenizers, $T_A$ and $T_B$, tokenize string $S$ into a $2$-gram and $3$-gram respectively. 
If $S$ is informative, ideally, it should not matter in which $n$-gram embedding space this information is stored. 
We view $N$-gram as a sampling method to shortlist potential byte sequences, from an exponential number of permutations, for the model to learn and determine its usefulness.

\paragraph{Tokenizer Agnosticism.} 
Hence, our goal is to modify the Engram Module such that it can be (re)used regardless of tokenizer choice.
In this work, our contributions are as follow:
\begin{enumerate}
    \item Define the objective of Tokenizer Agnosticsm with respect to the Engram module, i.e., its hashing component.
    \item Propose a straightforward replacement of XOR hashing with the general polynomial hashing, detailing its advantages and maintaining similar algorithmic efficiency.
    \item Investigate the potential trade-offs of modelling in a joint $N$-gram embedding space. From our experiments, we obtain comparable results that suggest no downsides.\footnote{We provide our training and data processing code, built on top of Lightning and HuggingFace, in the Code and Data Supplement.}
    \item We train a model with a different tokenizer on pretrained Engram embeddings for cross-tokenizer transfer, empirically showing that tokenizer agnosticism can be achieved.
\end{enumerate}

\begin{table*}[t]
\centering

\begin{tabular}{l|rrrrrrrr}
  \toprule
  \multicolumn{9}{c}{Example Text: ``This is Genghis Khan''}\\
  \midrule
  Tokenizer & \multicolumn{8}{c}{Tokens} \\
  \midrule
  \texttt{Mistral} & '<s>'& '\_This'& '\_is'& '\_Gen'& 'gh'& 'is'& '\_Khan'& '</s>’\\
  \texttt{SmolLM2} & 'This'& 'Ġis'& 'ĠGen'& 'gh'& 'is'& 'ĠKhan'& '<|endoftext|>’\\
  \texttt{cl100k\_base} & 'This'& 'Ġis'& 'ĠG'& 'eng'& 'his'& 'ĠKhan'& '<|endoftext|>’\\
  \bottomrule
\end{tabular}
  \caption{A simple example showing the possible differences from using different tokenizers: (i) order not guaranteed due to possible shifts (see \texttt{Mistral}/\texttt{SmolLM2}); (ii) $N$-grams with different partitions (see 'Genghis' in \texttt{SmolLM2}/\texttt{cl100k\_base}); (iii) tokenizer specific characteristics, e.g., space prefix, tokenizer indices, etc. For this example, we use the text representation of the tokens. These confounding factors complicates the mapping between the same $N$-grams from different tokenizers.}
  \label{tab:tokenizer_example}
\end{table*}

\section{Related Works}

\paragraph{Tokenization.} Besides common tokenization methods such as BPE and Unigram. SuperBPE \cite{liu2025superbpe} and BoundlessBPE \cite{schmidt2025boundless} are tokenization methods that include tokens across predefined delimiters, e.g., white spaces. These additional tokens selected can be viewed as token-level $N$-grams. These $N$-grams are preselected by the tokenizer algorithm, whereas for Engram module, the model learns which $N$-grams are useful during training.

\paragraph{Engram Module Adjacent.} 
There are other alternative embedding modules/layers \cite{google2025gemma, tseng2026l, sadhukhan2026stem} whose mechanisms are different from Deepseek's Engram. Our work focuses on removing the latter's tokenizer requirement. \citet{zheng2026lngram} proposes Lngram which learns discrete symbols from the latent space for embedding lookup; and not via tokens. It trades compute efficiency, as the lookup is reliant on the previous model block's output for model performance.

\paragraph{Hash Embeddings.} First proposed by \citet{svenstrup2017hash}, it shares similarity with DeepSeek's multihead embeddings and context-aware gating \cite{cheng2026conditionalmemoryscalablelookup}. Our work also uses a shared embedding space, with the main difference being hashing from byte sequences to achieve the byte-equivalence property.

\begin{figure*}
\centering
\begin{subfigure}[t]{0.48\textwidth}
    \includegraphics[width=\linewidth]{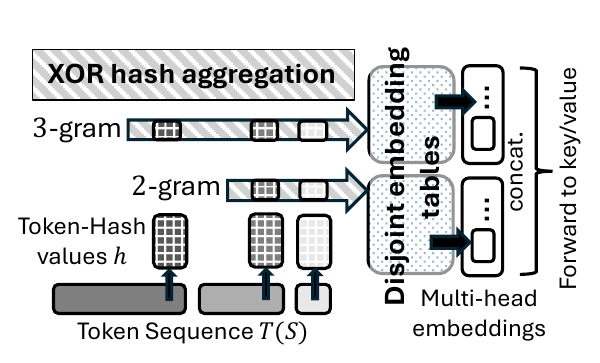}
    \caption{Original implementation by DeepSeek \cite{cheng2026conditionalmemoryscalablelookup}.}
    \label{fig:vis_1}
\end{subfigure}
\;\;\;\;\;\;
\begin{subfigure}[t]{0.4\textwidth}
    \includegraphics[width=\linewidth]{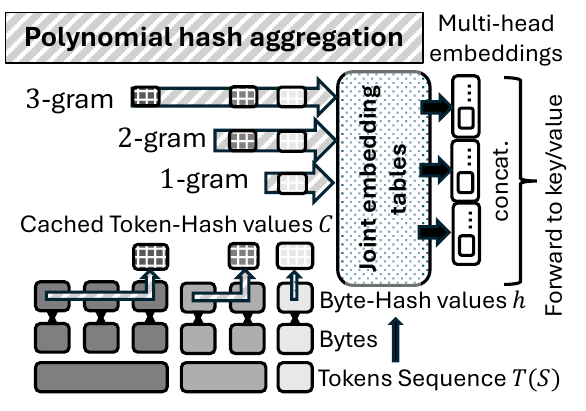}
    \caption{Ours: we hash from bytes, include $1$-gram and use shared joint embeddings.}
    \label{fig:vis_2}
\end{subfigure}
\caption{Visualization of the algorithmic and architectural changes between the original approach and our proposed approach. For our example, we set $N=3$, and we use a token sequence $T(S)$ of three tokens with different byte lengths.}
\label{fig:visualization}
\end{figure*}

\section{Hashing for Tokenizer-Agnosticism}

Extending the original tokenizer compression rules, e.g., such as capitalization and space prefix, we also account for special tokens and space prefixes between different tokenizers.
However, they are still likely to produce different token sequences. 
Even if they share some token vocabulary, there is no guarantee that the token sequences will be in the same order or that they will be partitioned using the same tokens.

\paragraph{Example.} In Table~\ref{tab:tokenizer_example}, we show (i) token sequences from \texttt{Mistral} and \texttt{SmolLM2} differ in token order, and (ii) token sequences from \texttt{SmolLM2} and \texttt{cl100k\_base}, ``Genghis'' are partitioned differently.

\subsection{Problem: Hashing Equivalence between Tokenizers}

We illustrate the problem using two different toy BPE tokenizers $T$ and their indexed token sets, only processing strings containing characters `a' and `b'.
\[
    T_A:\{\text{0: `aa', 1: `a', 2: `b'}\}
\]
\[
    T_B:\{\text{0: `a', 1: `b'}\}
\]

Given a string text $S = \text{``aabb''}$, the respective tokenizers will produce the following string token sequence.
\[
    T_A(S): (\text{`aa', `b', `b'})
\]
\[
    T_B(S): (\text{`a', `a', `b', 'b'})
\]

Likewise with actual tokenizers, it is not guaranteed that two different tokenizers will produce the same tokenized output.
We define $\|T(S)$ as the in-sequence concatenation of bytes from $T(S)$. It is evident that $S$ is recovered from $\|T_A(S)$ and $\|T_B(S)$, therefore they are \textit{byte equivalent}.
\begin{equation}
\|T_A(S) \equiv \|T_B(S) 
\label{eq:byte_equiv}
\end{equation}
A na\"ive approach to map byte-equivalent $N$-grams would involve comparing their bytes.
Byte equivalence ensures that we can map different token sequences of the same bytes to the same embedding index.
\[
Hash(T_A(S)) \equiv Hash(T_B(S)).
\]

Notice that for this example $S$, the 3-gram of $T_A$ is byte-equivalent to the 4-gram of $T_B$.
To achieve tokenizer-agnoticism, it is clear that we require byte-level information, and not only rely on token-level information, to ensure equivalence of $N$-grams from different tokenizers. However, this requirement itself presents further challenges:
\begin{enumerate}
    \item Time. Comparing bytes instead of integer tokens results in a more expensive computation as $|S| \geq |T(S)|$.
    \item Space. Na\"ively storing permutations will not work for large vocabulary size as it requires $O(|T|^N)$ space.
\end{enumerate}
An efficient method will involve addressing these concerns, in addition to ensuring byte-equivalent $N$-grams, using common strategies such as:
\begin{enumerate}[label=\roman*.]
    \item Caching hashes for each integer token to reduce repeated computations. 
    \item On-demand calculation of final $N$-gram hashes from the cached integer token hashes.
\end{enumerate}

\paragraph{Objective.} For any token sequences up to length $N$, we expect \textit{hash equivalence} between the different integer token sequences, from different tokenizers, when there is byte equivalence:
\begin{equation}
Hash(T_A(S)) \equiv Hash(T_B(S)) | \text{\;Eq. \ref{eq:byte_equiv}}.
\end{equation}
Note that hash collisions might occur, i.e., there is hash equivalence even when Eq.~\ref{eq:byte_equiv} is not true, simply due to chance.

\subsection{XOR-hashing}

In the original implementation, $N$-gram indices were derived by applying XOR to token hashes (see Fig.~\ref{fig:vis_1}). Given a mapping of token indices to hash $h$ and hyperparameters $base$ and $M$ (allocated Engram embedding table size),
\[
Hash_x(T(S)) = [\bigoplus^{|T(S)|-1}_{i=0} h(T(S)_{i})\cdot base_{i}] \mod M.
\]
This is insufficient to achieve hash equivalence for byte-equivalent $N$-grams. 

\paragraph{Counterexample 1.} Two byte-equivalent $N$-grams from different tokenizers might differ in the partitions. Given $S_1 = (`ab',`a')$ and $S_2 = (`a',`ba')$, with different token hashes $h$ and $M$ omitted, unless collision by chance,
\[
h(`ab') \cdot base_0 \oplus h(`a')\cdot base_1 \not\equiv h(`a') \cdot base_0 \oplus h(`ba')\cdot base_1.
\]

\paragraph{Counterexample 2.} Two byte-equivalent $N$-grams differ in $N$.
Given $S_1 = (`aba')$ and $S_2 = (`a',`ba')$, with different token hashes $h$ and $M$ omitted, unless collision by chance,
\[
h('aba')\cdot base_0 \not\equiv h(`a') \cdot base_0 \oplus h(`ba')\cdot base_1.
\]
For both examples, we get different hashes for byte-equivalent $N$-grams. A simple fix could involve removing the positional $base$ hyperparameters.
However, XOR's commutative property increases hash collisions:
\begin{enumerate}
    \item Information destruction. Given $S_1 = (`a', `b', `a')$, $S_2 = (`b', `a', `a')$, and a byte-hash map function $h$. A $h(`a') \oplus h(`a') = 0$, which makes it a hash of odd-count unique tokens or bytes, $Hash(S_1) \equiv Hash(S_2)$ when $S_1 \not\equiv S_2$.
    \item Lack of position information. Given $S_3 = (`a', `b', `c')$, $S_4 = (`c', `a', `b')$, the order independence results in $Hash(S_3) \equiv Hash(S_4)$ when $S_3 \not\equiv S_4$, affecting the byte anagrams produced by the previous scenario.
\end{enumerate}
Another possible approach involves mapping the tokens between two tokenizers.
However, it is unclear how these non-one-to-one hash embeddings will aggregate or disentangle during training.
Such methods are likely to require additional complex mechanics.

\subsection{Polynomial-hashing as an Efficient Alternative}
To achieve hash equivalence for byte-equivalent sequences, we need to accumulate hash information.
Polynomial-hashing \cite{carter1979universal, Bhattacharyya2025poly} is a common hashing approach that fits this criterion. 
Given string $S$, byte-hash map function $h$, and hyperparameter $base$ and $M$ (Engram embedding table size),
\begin{equation}
    Hash_{p}(S) = [\sum^{|S|-1}_{i=0} h(s_i) \cdot base^{|S|-i} ]\mod M .
\label{eq:polyhash_sum}
\end{equation}
Eq.~\ref{eq:polyhash_sum} can be rewritten as an equivalent streaming variant,
\begin{align}
\begin{split}
    Hash_{p}(S) &= [\dots(h(s_0)\cdot base 
    + h(s_1))\cdot base \dots \\
    &+ h(s_{|S|-1}))\cdot base] \mod M .
\end{split}
\label{eq:polyhash_stream}
\end{align}
However, as previously described, we have to cache the token hashes to efficiently calculate the final $N$-gram hash. 
We can do so by caching the hash of each token $t \in T$, $C(t)$ (Eq. \ref{eq:cache_token}), and $base$ exponents, $B(power)$ (Eq. \ref{eq:cache_base}).
\begin{equation}
    C(t) = Hash_{p}(t)
    \label{eq:cache_token}
\end{equation}
\begin{equation}
    B(power) = base^{power}
    \label{eq:cache_base}
\end{equation}
We can then modify Eq.~\ref{eq:polyhash_sum} to use Eq.~\ref{eq:cache_token} and Eq.~\ref{eq:cache_base},
\begin{align}
\begin{split}
    Hash_{p}(T(S)) =& [\sum_{i = 0}^{|T(S)|-1} C(T(S)_i) \\ &\cdot B(\sum_{j=i+1}^{|T(S)|-1} |T(S)_j|) ] \mod M.
\end{split}\label{eq:polyhash_sum_cache}
\end{align}
The equivalent streaming variant of Eq.~\ref{eq:polyhash_sum_cache} is more convenient as it removes the requirement to track remaining length,
\begin{align}
\begin{split}
    Hash_{p}(&T(S)) = [\dots(C(T(S)_0)\cdot B(|T(S)_1|) \\
    &+ C(T(S)_1)) \cdot B(|T(S)_2|) \dots \\
    &+ C(T(S)_{|T(S)|-1}))\cdot B(0)] \mod M .
\end{split}\label{eq:polyhash_stream_cache}
\end{align}
After initializing the cache, computing the polynomial hash of a token sequence $T(S)$ is $O(|T(S)|\cdot N)$ and can be computed using matrix operations, achieving the same complexity as original XOR-hash routine.

\paragraph{Example.} Given string $S=``abcd''$, where its token sequence $T(S)=(``ab'',``cd'')$, following Eq.~\ref{eq:polyhash_stream}, we get
\begin{align}
\begin{split}
   Hash_p(S) =[(((h(`a')\cdot base + h(`b')) \cdot base\\
    +h(`c')) \cdot base + h(`d')) \cdot base] \mod M.
\end{split}
\label{eq:example}
\end{align}
The following cached values for its tokens $T(S)$ are
\begin{align*}
    C(``ab'') &= (h(`a')\cdot base + h(`b')) \cdot base, \\
    C(``cd'') &= (h(`c')\cdot base + h(`d')) \cdot base.
\end{align*}
Rearranging Eq.~\ref{eq:example} shows that Eq.~\ref{eq:polyhash_stream} $\equiv$ Eq.~\ref{eq:polyhash_stream_cache},
\begin{align}
\begin{split}
    &Hash_p(S) = [((h(`a')\cdot base + h(`b')) \cdot base)\cdot base^2 \\
    &+ ((h(`c')\cdot base + h(`d')) \cdot base) \cdot base^0]\mod M \\
    &= [C(``ab'')\cdot B(|``cd''|) + C(``cd'')]\mod M \\
    &= Hash_p(T(S)).
\end{split} \nonumber
\end{align}
Simply substituting the hashing mechanism enables us to preserve similar algorithmic efficiency. 
We use the general polynomial hashing approach for this work.
Other specific hashing approaches, such as \citet{bernstein2005poly1305, degabriele2024sok}, may be applicable.

\subsection{Research Questions}

When comparing the different token sequences from different tokenizers, there are three key scenarios that will influence the training of tokenizer-agnostic Engram embeddings.

\paragraph{$N$-gram does not exist.} A $N$-gram from a tokenizer A might never be found in any token sequence from tokenizer B. This disparity can be largely attributed to different tokenization behaviours and vocabulary. Theoretically, when $N$ is large enough, there will be more chances to align byte-equivalent $N$-grams from different tokenizers. Additionally, most tokenizers preprocess texts similarly, e.g., splitting before space, and thus share similar points of partitions. Nevertheless, are there enough byte-equivalent $N$-grams between different tokenizers such that pretrained engram embeddings will be useful?

\paragraph{Joint embedding space for different $N$-grams.} Due to $N$-gram mismatch, different tokenizers might partition the same byte sequence into different numbers of tokens, e.g., a $3$-gram from a target tokenizer might be a $2$-gram of another reference tokenizer. The original implementation models $N$-grams in a disjoint manner, i.e., embedding spaces between different $N$ are disjoint. 
This means that even if the exact byte-equivalent $N$-gram had been learnt, due to the allocation of the embedding table, the information is inaccessible unless $N$ is similar for both tokenizers. 
A straightforward solution will involve removing the specific $N$-gram allocations and allow mixing between the $N$-grams of different $N$. 
This would imply that the Engram embeddings will lose useful $N$-related information, if any, and such information will be learnt upstream as the down-projection of concatenated $N$-gram embeddings remains unchanged. 
Will this $N$-gram mixing adversely affect the learning of Engram embeddings?

\paragraph{$1$-gram requirement.} The original implementation does not model 1-gram information as the backbone model is responsible for that.
Between different tokenizers, it is plausible that single tokens from one tokenizer will be partitioned into multiple tokens when using another tokenizer.
If we ignore 1-grams in the Engram embedding space, there could be information loss when transferring the Engram embeddings to a model with a different tokenizer.
To include 1-grams, there will be a small increase in parameters for the down-projection to key/values.
Since 1-gram hashes will be frequently accessed and updated during training, will they adversely affect the learning of Engram embeddings?

\begin{figure*}[t]
    \centering
\begin{subfigure}[t]{0.38\textwidth}
    \includegraphics[width=\linewidth]{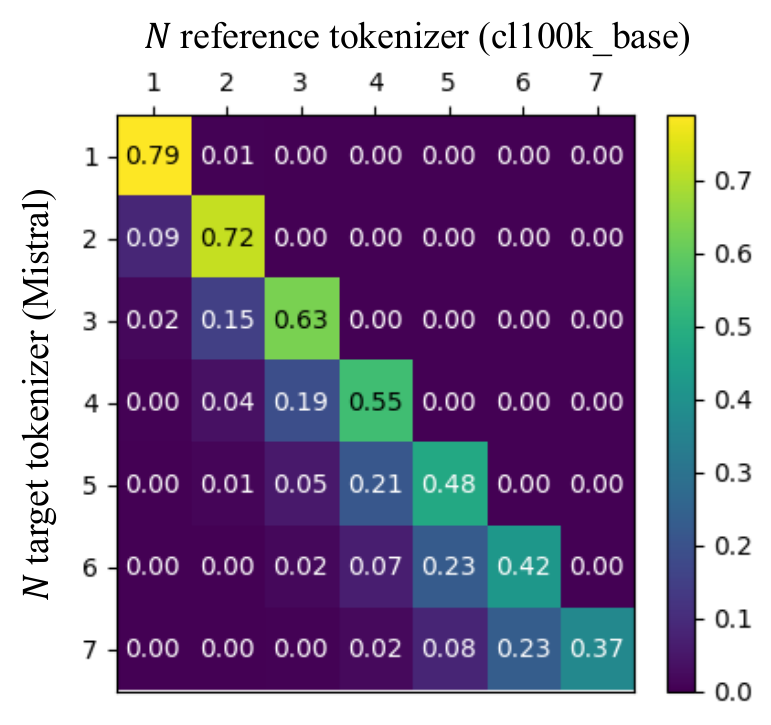}
    \caption{\% of unique $N$-grams from \texttt{Mistral} tokenizer found in \texttt{cl100k\_base} tokenizer}
    \label{fig:ngrams_ratio_a}
\end{subfigure}
\:\:\:\:\:\:\:\:\:\:
\begin{subfigure}[t]{0.38\textwidth}
    \includegraphics[width=\linewidth]{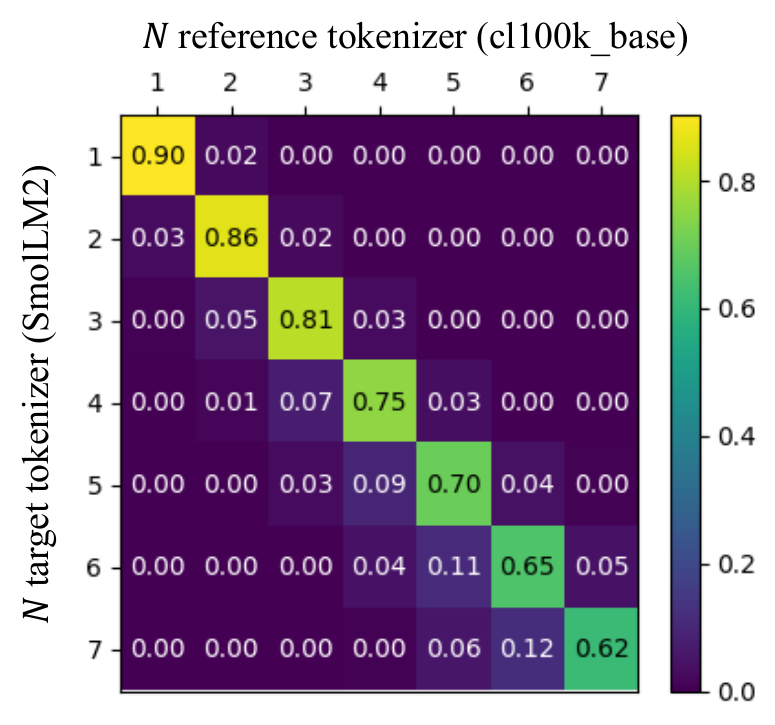}
    \caption{\% of unique $N$-grams from \texttt{SmolLM2} tokenizer found in \texttt{cl100k\_base} tokenizer}
    \label{fig:ngrams_ratio_b}
\end{subfigure}
\caption{Percentages of unique $N$-grams sampled on wikitext $val$ using different target tokenizers (\texttt{Mistral} and \texttt{SmolLM2}) that exists when using reference tokenizer (\texttt{cl100k\_base}). $N$-gram pairs, on the diagonal, might be partitioned differently.}
\label{fig:ngrams_ratio}
\end{figure*}

\section{Experiments}

\subsection{Backbone Language Models}

For our experiments, we pretrain from scratch multiple variants using these three decoder-only transformer backbones:\footnote{One run for each model variants on a 8xH200 GPU node, 4096 max length token packing, conv. kernel size 4 for Engram module.}
\begin{enumerate}
    \item \texttt{SmolLM2-1.7B} \cite{allal2025smollm2}. The underlying architecture is \texttt{Llama2} \cite{touvron2023llama2openfoundation}, with full SPDA attention blocks \cite{vaswani2017attentionneed}.
    \item \texttt{Qwen2-7B} \cite{yang2024qwen2technicalreport}. Uses Group Query Attention (GQA) \cite{ainslie2023gqa} instead of full attention.
    \item \texttt{Qwen3.5-0.8B-text-only}\footnote{principled-intelligence/Qwen3.5-0.8B-text-only} \cite{qwen3.5}. While this model can process both image and text modalities, we only utilize its 0.7B text-relevant parameters. It uses Gated Delta Net \cite{yang2025gated} blocks, at a ratio of 3:1 to transformer blocks.
\end{enumerate}
These models use mostly default settings from HuggingFace's repository, with changes to use other tokenizers and to add Engram modules.\footnote{Only two blocks: after first block and in middle of backbone.} 
We allocate $\sim$35\% of total parameters to the Engram module, similar to \citet{cheng2026conditionalmemoryscalablelookup}.\footnote{This parameter budget anchors our hyperparameters selection.}
During pretraining, we use AdamW \cite{loshchilov2018decoupled} optimizer with a cosine scheduler with warm up.

\subsection{Benchmark and Metrics.}
We use several popular benchmarks, from EleutherAI's \texttt{lm-evaluation-harness} \cite{eval-harness}, that are commonly-used for pretraining evaluation: ARC \cite{clark2018arc}, BoolQ \cite{clark2019boolq}, COPA \cite{roemmele2011choice}, HellaSwag \cite{zellers2019hellaswag}, LAMBADA \cite{paperno2016lambada, radford2019language}, PIQA \cite{bisk2020piqa}, SCIQ \cite{welbl2017sciq}, and Winogrande \cite{levesque2011winograd, sakaguchi2019winogrande}.
When possible, for these multiple-choice question benchmarks, we use length-normalized accuracy (\texttt{acc\_norm}) instead of accuracy (\texttt{acc}).\footnote{Select choice candidate with the highest log-prob sum. We can normalize for tokens length to reduce bias towards long answers.}
For aggregation across benchmarks, we report the \texttt{Mean} of all accuracy and accuracy norm values together.
Finally, as we will be comparing across tokenizers, we report \texttt{bits-per-byte} of \texttt{wikitext} benchmark \cite{merity2016pointer}.

\subsection{Byte-equivalent $N$-grams Analysis}

First, we examine the occurrences of $N$-grams, from the target tokenizer, 1) that do not exist in the reference tokenizer and 2) byte-equivalent $N$-grams with different $N$, i.e., found in both target and reference tokenizer. 
We select three different tokenizers, with \texttt{Mistral} (${\sim}$32$K$ size) \cite{jiang-etal-2023-mistral} and \texttt{SmolLM2} (${\sim}$49$K$ size) tokenizers as target tokenizers, and \texttt{cl100k\_base} (${\sim}$100$K$ size) \cite{openai_tiktoken} as the reference tokenizer.
We employ our target and reference tokenizers to tokenize wikitext $val$, consisting of 641 documents totaling $~$2M tokens, obtaining unique $N$-grams varying $N \in [1,7]$. 
We then compare the obtained unique $N$-gram sets, per document.
We expect some target $N$-grams to be missing from the reference $N$-grams. 
A reference tokenizer with a larger tokenizer vocabulary tends to compress better, which may result in fewer unique $N$-grams. 

\begin{table}[t]
    \centering
    \resizebox{.45\textwidth}{!}{%
    \begin{tabular}{l|ccccccc} \toprule
         \multicolumn{1}{c}{} & \multicolumn{7}{c}{$N$}\\ \cmidrule{2-8}
         Target & 1 & 2 & 3 & 4 & 5 & 6 & 7 \\ \cmidrule(lr){1-2}\cmidrule{2-8}
         \texttt{Mistral} & 0.80 & 0.81 & 0.80 & 0.78 & 0.75 & 0.74 & 0.70 \\
         \texttt{SmolLM2} & 0.92 & 0.91 & 0.89 & 0.86 & 0.86 & 0.85 & 0.80 \\\bottomrule
    \end{tabular}
    }
    \caption{\% of unique wikitext $val$ $N$-grams obtained from target tokenizer and can be found in the unique $N$-grams of reference tokenizer. Refer to Fig.~\ref{fig:ngrams_ratio} for the breakdown.}
    \label{tab:ngrams_no_exist}
\end{table}

\paragraph{Byte-equivalent $N$-grams exists across tokenizers.} From Table~\ref{tab:ngrams_no_exist}, observe that there is a healthy percentage of unique $N$-grams, sampled from wikitext $val$, from our target tokenizers and found in the reference tokenizer. 
If these byte-equivalent $N$-grams exist, this means that information stored in their respective Engram embeddings can be accessed despite the difference in tokenizers.

\paragraph{Byte-equivalent $N$-grams with different $N$.} From our detailed breakdown in Figure~\ref{fig:ngrams_ratio}, observe that when target $N$ increases, the proportion of byte-equivalent $N$-gram found in other reference $N$ increases substantially. This implies that without enforcing the byte-equivalence constraint across $N$, we decrease the possible avenues for cross-tokenizer information transfer via $N$-grams.

\begin{table*}[t]
\centering
\resizebox{0.98\textwidth}{!}{
\begin{tabular}{ccccccccccccc|cc}
\toprule
Experiment & Engram? & Compare &
  Arc$_c$ & Arc$_e$ & BoolQ & COPA & Hella. & LAMB. & PIQA & SCIQ & Wino &
  \textbf{Mean} $(\uparrow)$ &
  \textbf{bits/byte} $(\downarrow)$ \\\cmidrule(lr){1-14}
\multicolumn{1}{c}{Base 1.7B} &
  NO & Reference&  { 0.277} & { 0.534} & { 0.581} & { 0.670} & { 0.472} & { 0.482} & { 0.704} & { 0.758} & { 0.523} & { 0.556} & { 0.861} \\\cmidrule(lr){1-14}
\multirowcell{4}{Base 1.7B\\+ Engram 0.8B \\+ XOR} &
  \multirow{2}{*}{NO} & - & { 0.312} & { 0.578} & { 0.616} & { 0.720} & { 0.501} & { 0.495} & { 0.701} & { 0.798} & { 0.559} & { 0.587} & { 1.034} \\
 & & To Base (\%) & { 0.126} & { 0.082} &
\textbf{ 0.060} & { 0.075} & { 0.061} & { 0.027} &
  -0.004 & { 0.053} & { 0.069} & { 0.056} & { 0.201} \\\cmidrule(lr){2-14}
 &
  \multirow{2}{*}{YES} & - & { 0.317} & { 0.614} & { 0.598} & { 0.700} & { 0.546} & { 0.558} & { 0.724} & { 0.825} & { 0.567} & { 0.605} & { 0.919} \\
 & & To Base (\%) & { 0.144} & { 0.150} & { 0.029} & { 0.045} & { 0.157} & { 0.158} & { 0.028} & { 0.088} &
\textbf{ 0.084} & { 0.090} &
  \textbf{ 0.067} \\\cmidrule(lr){1-14}
\multirowcell{4}{Base 1.7B \\+ Engram 0.8B \\+ Poly} &
  \multirow{2}{*}{NO} & - & { 0.318} & { 0.604} & { 0.530} & { 0.730} & { 0.533} & { 0.513} & { 0.714} & { 0.821} & { 0.564} & { 0.592} & { 1.025} \\
 & & To Base (\%) & { 0.148} & { 0.131} &
  -0.088 & { 0.090} & { 0.129} & { 0.064} & { 0.014} & { 0.083} & { 0.078} & { 0.065} & { 0.190} \\ \cmidrule(lr){2-14}
 &
  \multirow{2}{*}{YES} & - & { 0.328} & { 0.629} & { 0.556} & { 0.760} & { 0.559} & { 0.566} & { 0.727} & { 0.827} & { 0.560} & { 0.612} & { 0.925} \\ 
 &  & To Base (\%) & \textbf{ 0.184} & \textbf{ 0.178} & -0.043 & \textbf{ 0.134} & \textbf{ 0.184} & \textbf{ 0.174} & \textbf{ 0.033} & \textbf{ 0.091} & { 0.071} & \textbf{ 0.102} & { 0.074} \\
 \bottomrule
\end{tabular}}
\caption{We pretrain from scratch a reference base model (\texttt{SmolLM2-1.7B}), and two Engram models differing in hashing mechanism: original XOR hashing and the proposed Polynomial hashing substitution. We train the models using 32B \texttt{SmolLM2} tokens from \texttt{dclm-dedup} corpus. The models using XOR and Polynomial have similar training hyperparameters. Results here shows that using Polynomial hashing does not negatively affect training and have comparable results to XOR-based hashing. For the models with Engram, we additionally evaluate its backbone without the Engram module to ascertain its value-add. Engram? NO indicates that we switch off the Engram parameters. Standard error for all accuracy-metrics benchmarks $\leq 0.015$.}
\label{tab:pretraining_1}
\end{table*}

\subsection{Comparing XOR and Poly. Hashing Engrams}

For this experiment in this subsection, we wish to determine whether the requirements of 1) $1$-gram and 2) joint embedding spaces will affect the pretraining of Engram modules. We train three models using \texttt{SmolLM2-1.7B} as the backbone with its native tokenizer, on 32B tokens from \texttt{dclm-dedup} \cite{tokpanov-etal-2024-zyphra}:
\begin{enumerate}[label=\alph*.]
    \item \texttt{Base-1.7B}. As the main comparisons of our subsequent models are along various dimensions, this model serves as a \emph{common} point of reference and normalizes the results for easier comparison. The difference between the subsequent models and the base models highlights the value add from the additional Engram module, giving a better perspective compared to a direct comparison.
    \item Additional 0.8B Engram module with hyperparameter $N=3$, 1M indices of size 192 per block.
    \begin{enumerate}[label=\roman*.]
        \item Using the original XOR hashing, we adapt DeepSeek's demonstration code\footnote{Original demo code: github.com/deepseek-ai/Engram} for our training infrastructure.
        \item Using our proposed polynomial hashing, there are three main changes. First, we directly substitute the hashing algorithm. Second, we remove the disjoint embedding spaces by using the same $base$ positional hyperparameter for all $N$. Lastly, we include $N=1$ hash indices per token. To reduce the impact of $N=1$ at the start of training, we scale its embedding values by $0.1$. Note that as training progress, if $N=1$ is indeed informative, this scaling will be made redundant.
    \end{enumerate}
\end{enumerate}

\begin{table*}[t]
\centering
\resizebox{0.98\textwidth}{!}{
\begin{tabular}{ccccccccccccc|cc}
\toprule
Experiment & Engram? & Compare &
  Arc$_c$ & Arc$_e$ & BoolQ & COPA & Hella. & LAMB. & PIQA & SCIQ & Wino &
  \textbf{Mean} $(\uparrow)$ &
  \textbf{bits/byte} $(\downarrow)$ \\\cmidrule(lr){1-14}
\multirowcell{2}{Model $\textbf{A}$ 7B \\+ Engram 2B}  & NO    & - & { 0.330} & { 0.617} & 0.540 & 0.780 & { 0.584} & { 0.551} & { 0.728} & { 0.837} & { 0.579} & { 0.616} & { 0.779} \\
\cmidrule(lr){2-14}
                                            & YES   & - & { 0.346} & { 0.660} & 0.594 & 0.770 &{ 0.627} &{ 0.611} &{ 0.757} &{ 0.858} &{ 0.590} &{ 0.646} &{ 0.708} \\
\cmidrule(lr){1-14}
\multicolumn{1}{c}{Model $\textbf{B}$ 0.7B}            & NO    & Reference & { 0.272} & { 0.539} & 0.553 & 0.700 &{ 0.458} &{ 0.499} &{ 0.690} &{ 0.772} &{ 0.537} &{ 0.560} & { 0.821} \\\cmidrule(lr){1-14}
\multirowcell{4}{Model $\textbf{B}$ 0.7B \\+ Pretrained Model $\textbf{A}$'s \\Engram Emb. 1.5B} 
& \multirow{2}{*}{NO}                               & - & { 0.256} &{ 0.509} & 0.571 & 0.700 &{ 0.437} &{ 0.356} &{ 0.686} &{ 0.710} &{ 0.525} &{ 0.528} &{ 0.905} \\
 &                                                  & To B (\%) & -0.059 & -0.056 & 0.033 & 0.000 & -0.046 & -0.287 & -0.006 & -0.080 & -0.022 & -0.063 &{ 0.102} \\\cmidrule(lr){2-14}
 & \multirow{2}{*}{YES}                             & - & { 0.27} & { 0.545} & 0.611 & 0.750 &{ 0.476} &{ 0.514} &{ 0.703} &{ 0.800} &{ 0.537} &{ 0.584} &{ 0.811} \\
 &                                                  & To B (\%) & \textbf{-0.007} & \textbf{ 0.011} & \textbf{0.105} & \textbf{0.071} & \textbf{ 0.039} & \textbf{ 0.030} &\textbf{ 0.019} &\textbf{ 0.036} & \textbf{0.000} &\textbf{ 0.038} & \textbf{-0.012}\\
 \bottomrule
\end{tabular}}
\caption{We pretrain from scratch a \texttt{Qwen2-7B} (Model $\textbf{A}$), with an 2B Engram module ($N=3$), on 150B tokens from \texttt{dclm-dedup}, tokenized using \texttt{cl100k\_base} tokenizer. After training Model $\textbf{A}$ (\texttt{Qwen3.5-0.8B-text-only}), we use the pretrained Engram embeddings as frozen parameters, connected via a ($N$ = 7) module, to pretrain Model $\textbf{B}$. Model $\textbf{B}$ was trained on 100B \texttt{SmolLM2} tokens from \texttt{dclm-dedup}. Variants of Model $\textbf{B}$ uses similar hyperparameters. Results shows that Model $\textbf{B}$ can use Model $\textbf{A}$'s pretrained Engram embeddings. Standard error for all accuracy-metrics benchmarks $\leq 0.015$.}
\label{tab:pretraining_2}
\end{table*}

We compare the inter-hashing results between the different hashing approaches and intra-hashing results where we isolate the contribution of the Engram module. If the additional Engram module is redundant, we expect no performance difference when the Engram module is deactivated, as the model is wholly reliant on the backbone model.

\paragraph{Results.} As reported in Table~\ref{tab:pretraining_1}, for our first comparison, we examine the difference between models with different hashing approaches.
With the exception of BoolQ\footnote{Latter experiments suggests this is resolved via scaling tokens.}, the results of polynomial hashing, from other benchmarks, are comparable to that of XOR hashing.
This suggests that modifications for $1$-gram and joint-embedding spaces did not degrade pretraining performance. 
When we compare the results between active and inactive engram modules, within the same Engram models, we observe large improvements in mean and bits-per-byte which suggests that there are contributions from the additional Engram modules.

\paragraph{Discussion.} From this empirical experiment, we can conclude that Engram modules are useful in both Engram models. The inclusion of $1$-gram does not seem to negatively impact training, which we will again verify in our subsequent ablation.
The key benefit of modelling disjoint embedding spaces allows us to designate the proportion of embedding parameters to specific $N$, controlling the importance of specific $N$.
A larger proportion assigned will decrease the chances of hash collision for that specific $N$-gram. 
Our experiment assumed equal proportion.
The context-aware SPDA mechanism, in the Engram module, serves to gate these collisions.
Hence, collisions across $N$ may not matter as much or are mitigated by using the complete embedding table.
The similarity of the results suggests that the useful information lies at the byte-level and by happenstance in the $N$-gram space, and the main role of $N$-grams is to sample and shortlist potential useful byte sequences.

\begin{table*}[t]
\centering
\resizebox{0.98\textwidth}{!}{
\begin{tabular}{cccccccccccc|c}
\toprule
Experiment & Engram? &
  Arc$_c$ & Arc$_e$ & BoolQ & COPA & Hella. & LAMB. & PIQA & SCIQ & Wino &
  \textbf{Mean} $(\uparrow)$ &
  \textbf{bits/byte} $(\downarrow)$ \\\cmidrule(lr){1-13}
\multirow{2}{*}{Model $\textbf{C}$ (N=1)} & NO & 0.253 & { 0.493} & 0.615 & { 0.660} &{ 0.432} & 0.455 &{ 0.687} & 0.742 &{ 0.518} & 0.542 & { 0.851} \\\cmidrule(lr){2-13}
 & YES & 0.254 &{ 0.491} & 0.615 & { 0.670} & 0.431 & 0.457 &{ 0.691} & 0.743 &{ 0.514} & 0.544 &{ 0.851} \\\cmidrule(lr){1-13}
\multicolumn{2}{r}{Difference YES/NO (\%)} & 0.004& -0.004 & 0.000 & \textbf{ 0.015} & -0.002 & 0.004 &{ 0.006} & 0.001 & -0.008 & 0.003 &   { 0.000} 
\\\cmidrule(lr){1-13}
\multirow{2}{*}{Model $\textbf{B}$ (N=7)} & NO &{0.244} &{ 0.495} & 0.596 & { 0.710} &{ 0.426} & 0.402 &{ 0.687}& 0.700 &{ 0.515} & 0.533 & { 0.872} 
\\\cmidrule(lr){2-13} & YES &0.252 & 0.501 & 0.615 & { 0.680} &{ 0.438} & 0.455 &{ 0.685 }& 0.739 &{ 0.530} & 0.546 &{ 0.845} 
\\\cmidrule(lr){1-13}
\multicolumn{2}{r}{Difference YES/NO (\%)} &\textbf{0.033} & \textbf{ 0.012} & \textbf{0.032} & -0.042 & \textbf{ 0.028} & \textbf{0.132} & \textbf{-0.003} & \textbf{0.056} &\textbf{ 0.029} & \textbf{0.025} & \textbf{-0.031}\\
 \bottomrule
\end{tabular}}
\caption{Ablating results from Table~\ref{tab:pretraining_2}, we have Engram Model $\textbf{C}$, of similar architecture and tokenizer to Engram Model $\textbf{B}$, with the exception of setting $N=1$. Model $\textbf{C}$ is pretrained on 32B tokens and compared to an equivalent checkpoint of Model $\textbf{B}$. Results from this table show that there is little contribution from the Engram module when $N=1$, implying that the improvements came from using $N$-grams of $N>1$. Standard error for all accuracy-metrics benchmarks $\leq 0.015$.}
\label{tab:pretraining_3}
\end{table*}

\subsection{Training Tokenizer-agnostic Engrams}

Our reference tokenizer is \texttt{cl100k\_base}, and our target tokenizer is \texttt{SmolLM2}.
This experiment examines whether our proposed hashing approach allows a model with a target tokenizer to use Engram embeddings trained from a different reference tokenizer.
We pretrain Model $\textbf{A}$: a \texttt{Qwen2-7B} model with a 2B Engram module ($N=3$, 2M indices of size 192 per block) on 150B reference tokens from \texttt{dclm-dedup}. Then, we pretrain two variants of a smaller Model $\textbf{B}$ based on \texttt{Qwen3.5-0.8B-text-only} backbone: 
\begin{enumerate}[label=\alph*.]
    \item No Engram module, serving as the point of reference.
    \item We attach an Engram module ($N=7$)\footnote{We select a high $N$ to capture as many $N$-grams from the reference tokenizer. It is not necessarily advantageous to use a high $N$ due to hash collisions in the embedding space.}, that uses the previously pretrained engram embeddings as frozen parameters. Note that the other $\sim$15M parameters in the Engram module are trainable.
\end{enumerate}
Both variants of Model $\textbf{B}$ are trained on 100B target tokens from \texttt{dclm-dedup} in the same order flow. Our previous analysis show that there are many byte-equivalent $N$-grams across tokenizers. If these common $N$-grams are meaningful lookups to the pretrained Engram embeddings, we expect to observe an improvement in our selected benchmarks, indicating the value add of the additional parameters.

\paragraph{Results.} Between the Engram module activations and the reference model, in Table~\ref{tab:pretraining_2}, we observe that Model $\textbf{B}$+Engram shows improvements in most benchmarks, notably in BoolQ (+10\%) and COPA (+7\%). Without activating the Engram module, benchmark performance decreases, suggesting that the common $N$-grams are meaningful and cross-tokenizer lookups on pretrained Engram embeddings are successful, evidence of tokenizer agnosticism.

\subsection{Ablation Study}

Since our proposed method includes $1$-grams, could the model have relied on $1$-grams to achieve better performance?
We introduce another Engram Model $\textbf{C}$, similar to Engram Model $\textbf{B}$, except that we set $N=1$, essentially allowing the model to access only byte-equivalent $1$-grams found in the \texttt{SmolLM2} tokenizer. 
We pretrain Engram Model $\textbf{C}$ on 32B, on the same data flow as the corresponding checkpoint of the Engram Model $\textbf{B}$.\footnote{We intially pretrain Model $\textbf{A}$, $\textbf{B}$, and $\textbf{C}$ to 32B tokens, before deciding to scale Model $\textbf{A}$ and $\textbf{B}$ to 100B tokens.}

\begin{figure}
    \centering
    \includegraphics[width=\linewidth]{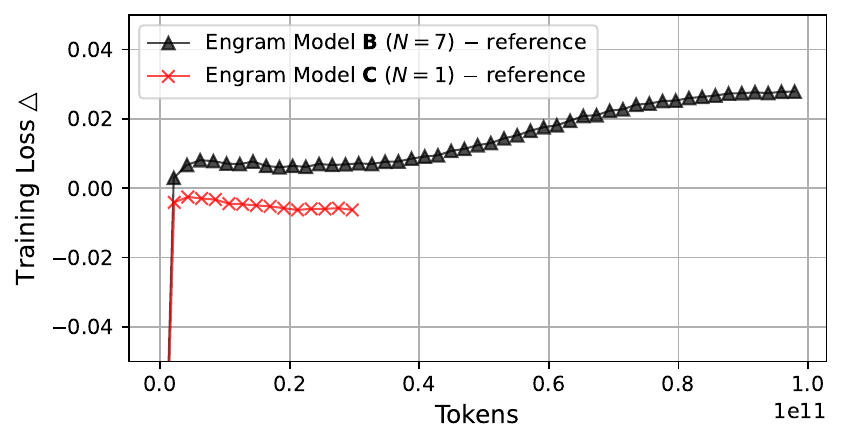}
    \caption{Plots of difference in training loss between Engram Models with pretrained Engram embeddings and reference model, against the 100B token training run. Each plot indicates an additional 2B tokens.}
    \label{fig:loss_curve}
\end{figure}

\paragraph{Results.} With the results in Table~\ref{tab:pretraining_3}, we can see that for Model $\textbf{C}$, there is only a marginal difference between activating and deactivating the engram module, compared to the more substantial differences from Model $\textbf{B}$.
This indifference suggests that for Model $\textbf{C}$, its Engram module was not beneficial. 
We further examine the training loss curves in Figure~\ref{fig:loss_curve}, which shows that the additional engram parameters, for Model $\mathbf{C}$, were counterproductive.
This rules out that the improvements were from $1$-grams, and that $(N>1)$-grams had successful lookups to the pretrained Engram embeddings.
Comparing between Engram module activation modes, this evaluation methodology shows that having more parameters does not necessarily lead to better performance.

\section{Conclusion and Future Work}
Reusing, modifying, and improving pretrained weights constitute an important and prevalent direction for cost-effective academic research.
We highlight the substitutability of byte-level and token-level $N$-gram information, to remove the tight tokenizer coupling from Engram module.
To achieve this, we propose using the general polynomial hashing approach to attain hash equivalence for byte-equivalent token sequences.
This requires us to modify the Engram module to accept 1-gram information and to model $N$-grams in a shared embedding space.
Our experiments show that these modifications are reasonable and enables effective cross-tokenizer transfer.
Possible future work directions involve scaling Engram in a tokenizer-agnostic manner. Beyond the dominant English language, it will also be interesting to investigate the multilingual capability of its hashed embeddings.

\bibliography{aaai2027}

@misc{openai_tiktoken,
  author = {OpenAI},
  title = {tiktoken: Fast BPE tokenizer for use with OpenAI's models},
  url = {https://github.com/openai/tiktoken},
  year = {2022}
}

@misc{eval-harness,
  author   = {Gao, Leo and Tow, Jonathan and Abbasi, Baber and Biderman, Stella and Black, Sid and DiPofi, Anthony and Foster, Charles and Golding, Laurence and Hsu, Jeffrey and Le Noac'h, Alain and Li, Haonan and McDonell, Kyle and Muennighoff, Niklas and Ociepa, Chris and Phang, Jason and Reynolds, Laria and Schoelkopf, Hailey and Skowron, Aviya and Sutawika, Lintang and Tang, Eric and Thite, Anish and Wang, Ben and Wang, Kevin and Zou, Andy},
  title = {The Language Model Evaluation Harness},
  month = 07,
  year = 2024,
  publisher = {Zenodo},
  version  = {v0.4.3},
  doi  = {10.5281/zenodo.12608602},
  url  = {https://zenodo.org/records/12608602}
}

@article{clark2018arc,
  author = {Peter Clark  and Isaac Cowhey and Oren Etzioni and Tushar Khot and
 Ashish Sabharwal and Carissa Schoenick and Oyvind Tafjord},
  title = {Think you have Solved Question Answering? Try ARC, the AI2 Reasoning Challenge},
  journal   = {arXiv:1803.05457v1},
  year  = {2018},
}

@inproceedings{zellers2019hellaswag,
 title={HellaSwag: Can a Machine Really Finish Your Sentence?},
 author={Zellers, Rowan and Holtzman, Ari and Bisk, Yonatan and Farhadi, Ali and Choi, Yejin},
 booktitle ={Proceedings of the 57th Annual Meeting of the Association for Computational Linguistics},
 year={2019}
}

@inproceedings{bisk2020piqa,
  author = {Yonatan Bisk and Rowan Zellers and Ronan Le Bras and Jianfeng Gao and Yejin Choi},
  title = {PIQA: Reasoning about Physical Commonsense in
   Natural Language},
  booktitle = {Thirty-Fourth AAAI Conference on
   Artificial Intelligence},
  year = {2020}
}

@article{welbl2017sciq,
 title={Crowdsourcing Multiple Choice Science Questions},
 author={Johannes Welbl, Nelson F. Liu, Matt Gardner},
 year={2017},
 journal={arXiv:1707.06209v1}
}

@inproceedings{clark2019boolq,
  title = {BoolQ: Exploring the Surprising Difficulty of Natural Yes/No Questions},
  author = {Christopher Clark and Kenton Lee and Ming-Wei Chang and Tom Kwiatkowski and Michael Collins and Kristina Toutanova},
  booktitle = {NAACL},
  year = {2019}
}

@inproceedings{paperno2016lambada,
    title = "The {LAMBADA} dataset: Word prediction requiring a broad discourse context",
    author = "Paperno, Denis  and
      Kruszewski, Germ{\'a}n  and
      Lazaridou, Angeliki  and
      Pham, Ngoc Quan  and
      Bernardi, Raffaella  and
      Pezzelle, Sandro  and
      Baroni, Marco  and
      Boleda, Gemma  and
      Fern{\'a}ndez, Raquel",
    
    booktitle = "Proceedings of the 54th Annual Meeting of the Association for Computational Linguistics (Volume 1: Long Papers)",
    year = "2016"
}

@article{merity2016pointer,
    title={Pointer Sentinel Mixture Models},
    author={Stephen Merity and Caiming Xiong and James Bradbury and Richard Socher},
    year={2016},
    journal={arXiv:1609.07843}
}

@article{sakaguchi2019winogrande,
    title={WinoGrande: An Adversarial Winograd Schema Challenge at Scale},
    author={Sakaguchi, Keisuke and Bras, Ronan Le and Bhagavatula, Chandra and Choi, Yejin},
    journal={arXiv preprint arXiv:1907.10641},
    year={2019}
}

@inproceedings{roemmele2011choice,
   title={Choice of plausible alternatives: An evaluation of commonsense causal reasoning},
   author={Roemmele, Melissa and Bejan, Cosmin Adrian and Gordon, Andrew S.},
   booktitle={2011 AAAI Spring Symposium Series},
   year={2011}
 }

@inproceedings{levesque2011winograd,
   title={The {W}inograd schema challenge},
   author={Levesque, Hector J and Davis, Ernest and Morgenstern, Leora},
   booktitle={{AAAI} Spring Symposium: Logical Formalizations of Commonsense Reasoning},
   volume={46},
   pages={47},
   year={2011}
 }

@article{vaswani2017attentionneed,
      title={Attention Is All You Need}, 
      author={Ashish Vaswani and Noam Shazeer and Niki Parmar and Jakob Uszkoreit and Llion Jones and Aidan N. Gomez and Lukasz Kaiser and Illia Polosukhin},
      journal={arXiv preprint arXiv:1706.03762},
      year={2017},
}

@article{ainslie2023gqa,
      title={GQA: Training Generalized Multi-Query Transformer Models from Multi-Head Checkpoints}, 
      author={Joshua Ainslie and James Lee-Thorp and Michiel de Jong and Yury Zemlyanskiy and Federico Lebrón and Sumit Sanghai},
      year={2023},
journal={arXiv preprint arXiv:2305.13245},
}

@article{radford2019language,
  title={Language Models are Unsupervised Multitask Learners},
  author={Radford, Alec and Wu, Jeff and Child, Rewon and Luan, David and Amodei, Dario and Sutskever, Ilya},
  year={2019},
  url={https://d4mucfpksywv.cloudfront.net/better-language-models/language_models_are_unsupervised_multitask_learners.pdf}
}

@article{jiang-etal-2023-mistral,
  title={{Mistral 7B}},
  author={Jiang, Albert Q and Sablayrolles, Alexandre and Mensch, Arthur and Bamford, Chris and Chaplot, Devendra Singh and Casas, Diego de las and Bressand, Florian and Lengyel, Gianna and Lample, Guillaume and Saulnier, Lucile and others},
  journal={arXiv preprint arXiv:2310.06825},
  year={2023}
}

@article{allal2025smollm2,
      title={SmolLM2: When Smol Goes Big -- Data-Centric Training of a Small Language Model}, 
      author={Loubna Ben Allal and Anton Lozhkov and Elie Bakouch and Gabriel Martín Blázquez and Guilherme Penedo and Lewis Tunstall and Andrés Marafioti and Hynek Kydlíček and Agustín Piqueres Lajarín and Vaibhav Srivastav and Joshua Lochner and Caleb Fahlgren and Xuan-Son Nguyen and Clémentine Fourrier and Ben Burtenshaw and Hugo Larcher and Haojun Zhao and Cyril Zakka and Mathieu Morlon and Colin Raffel and Leandro von Werra and Thomas Wolf},
      year={2025},
      journal={arXiv preprint arXiv:2502.02737},
}

@misc{qwen3.5,
    title  = {{Qwen3.5}: Towards Native Multimodal Agents},
    author = {{Qwen Team}},
    month  = {February},
    year   = {2026},
    url    = {https://qwen.ai/blog?id=qwen3.5}
}

@inproceedings{
yang2025gated,
title={Gated Delta Networks: Improving Mamba2 with Delta Rule},
author={Songlin Yang and Jan Kautz and Ali Hatamizadeh},
booktitle={The Thirteenth International Conference on Learning Representations},
year={2025},
url={https://openreview.net/forum?id=r8H7xhYPwz}
}

@misc{yang2024qwen2technicalreport,
      title={Qwen2 Technical Report}, 
      author={An Yang and Baosong Yang and Binyuan Hui and Bo Zheng and Bowen Yu and Chang Zhou and Chengpeng Li and Chengyuan Li and Dayiheng Liu and Fei Huang and Guanting Dong and Haoran Wei and Huan Lin and Jialong Tang and Jialin Wang and Jian Yang and Jianhong Tu and Jianwei Zhang and Jianxin Ma and Jianxin Yang and Jin Xu and Jingren Zhou and Jinze Bai and Jinzheng He and Junyang Lin and Kai Dang and Keming Lu and Keqin Chen and Kexin Yang and Mei Li and Mingfeng Xue and Na Ni and Pei Zhang and Peng Wang and Ru Peng and Rui Men and Ruize Gao and Runji Lin and Shijie Wang and Shuai Bai and Sinan Tan and Tianhang Zhu and Tianhao Li and Tianyu Liu and Wenbin Ge and Xiaodong Deng and Xiaohuan Zhou and Xingzhang Ren and Xinyu Zhang and Xipin Wei and Xuancheng Ren and Xuejing Liu and Yang Fan and Yang Yao and Yichang Zhang and Yu Wan and Yunfei Chu and Yuqiong Liu and Zeyu Cui and Zhenru Zhang and Zhifang Guo and Zhihao Fan},
      year={2024},
      journal={arXiv preprint arXiv:2407.10671},
}

@misc{touvron2023llama2openfoundation,
      title={Llama 2: Open Foundation and Fine-Tuned Chat Models}, 
      author={Hugo Touvron and Louis Martin and Kevin Stone and Peter Albert and Amjad Almahairi and Yasmine Babaei and Nikolay Bashlykov and Soumya Batra and Prajjwal Bhargava and Shruti Bhosale and Dan Bikel and Lukas Blecher and Cristian Canton Ferrer and Moya Chen and Guillem Cucurull and David Esiobu and Jude Fernandes and Jeremy Fu and Wenyin Fu and Brian Fuller and Cynthia Gao and Vedanuj Goswami and Naman Goyal and Anthony Hartshorn and Saghar Hosseini and Rui Hou and Hakan Inan and Marcin Kardas and Viktor Kerkez and Madian Khabsa and Isabel Kloumann and Artem Korenev and Punit Singh Koura and Marie-Anne Lachaux and Thibaut Lavril and Jenya Lee and Diana Liskovich and Yinghai Lu and Yuning Mao and Xavier Martinet and Todor Mihaylov and Pushkar Mishra and Igor Molybog and Yixin Nie and Andrew Poulton and Jeremy Reizenstein and Rashi Rungta and Kalyan Saladi and Alan Schelten and Ruan Silva and Eric Michael Smith and Ranjan Subramanian and Xiaoqing Ellen Tan and Binh Tang and Ross Taylor and Adina Williams and Jian Xiang Kuan and Puxin Xu and Zheng Yan and Iliyan Zarov and Yuchen Zhang and Angela Fan and Melanie Kambadur and Sharan Narang and Aurelien Rodriguez and Robert Stojnic and Sergey Edunov and Thomas Scialom},
      year={2023},
      journal={arXiv preprint arXiv:2307.09288},
}

@misc{tokpanov-etal-2024-zyphra,
    author = {Yury Tokpanov and Paolo Glorioso and Ayush Dattagupta and Vibhu Jawa and Ryan Wolf and Vikranth Jeyakumar and Arham Mehta and Quentin Anthony and Beren Millidge},
    title = {Building {Zyda-2}, a 5 {Trillion} {Token} {High-Quality} {Dataset}, with {NVIDIA} {NeMo} {Curator}},
    url = {https://www.zyphra.com/post/building-zyda-2},
    year = {2024},
}

@article{gage-1994-new,
author = {Gage, Philip},
title = {A new algorithm for data compression},
year = {1994},
issue_date = {Feb. 1994},
publisher = {R \& D Publications, Inc.},
address = {USA},
volume = {12},
number = {2},
issn = {0898-9788},
journal = {C Users J.},
month = {feb},
pages = {23–38},
numpages = {16},
url = {https://dl.acm.org/doi/10.5555/177910.177914}
}

@inproceedings{kudo-2018-subword,
  title={Subword regularization: Improving neural network translation models with multiple subword candidates},
  author={Kudo, Taku},
  booktitle={Proceedings of the 56th Annual Meeting of the Association for Computational Linguistics (Volume 1: Long Papers)},
  pages={66--75},
  year={2018}
}

@inproceedings{kudo-richardson-2018-sentencepiece,
    title = "{S}entence{P}iece: A simple and language independent subword tokenizer and detokenizer for Neural Text Processing",
    author = "Kudo, Taku  and
      Richardson, John",
    editor = "Blanco, Eduardo  and
      Lu, Wei",
    booktitle = "Proceedings of the 2018 Conference on Empirical Methods in Natural Language Processing: System Demonstrations",
    month = nov,
    year = "2018",
    address = "Brussels, Belgium",
    publisher = "Association for Computational Linguistics",
    url = "https://aclanthology.org/D18-2012",
    doi = "10.18653/v1/D18-2012",
    pages = "66--71",
}

@inproceedings{sennrich-etal-2016-neural,
    title = "Neural Machine Translation of Rare Words with Subword Units",
    author = "Sennrich, Rico  and
      Haddow, Barry  and
      Birch, Alexandra",
    editor = "Erk, Katrin  and
      Smith, Noah A.",
    booktitle = "Proceedings of the 54th Annual Meeting of the Association for Computational Linguistics (Volume 1: Long Papers)",
    month = aug,
    year = "2016",
    address = "Berlin, Germany",
    publisher = "Association for Computational Linguistics",
    url = "https://aclanthology.org/P16-1162",
    doi = "10.18653/v1/P16-1162",
    pages = "1715--1725",
}

@inproceedings{schmidt2025boundless,
    title={Boundless Byte Pair Encoding: Breaking the Pre-tokenization Barrier},
    author={Craig W Schmidt and Varshini Reddy and Chris Tanner and Yuval Pinter},
    booktitle={Second Conference on Language Modeling},
    year={2025},
    url={https://openreview.net/forum?id=oPAjXGV8qQ}
}

@inproceedings{liu2025superbpe,
    title={Super{BPE}: Space Travel for Language Models},
    author={Alisa Liu and Jonathan Hayase and Valentin Hofmann and Sewoong Oh and Noah A. Smith and Yejin Choi},
    booktitle={Tokenization Workshop},
    year={2025},
    url={https://openreview.net/forum?id=LwTWkSXIpt}
}

@misc{cheng2026conditionalmemoryscalablelookup,
      title={Conditional Memory via Scalable Lookup: A New Axis of Sparsity for Large Language Models}, 
      author={Xin Cheng and Rui Tian and Wangding Zeng and Damai Dai and Qinyu Chen and Bingxuan Wang and Zhenda Xie and Kezhao Huang and Xingkai Yu and Chengqi Deng and Shangyan Zhou and Chenggang Zhao and Zhewen Hao and Yukun Li and Han Zhang and Zhengyan Zhang and Yixu Wei and M. Y Xu and Huishuai Zhang and Dongyan Zhao and Wenfeng Liang},
      year={2026},
      journal={arXiv preprint arXiv:2601.07372}
}

@inproceedings{tao2024scaling,
    title={Scaling Laws with Vocabulary: Larger Models Deserve Larger Vocabularies},
    author={Chaofan Tao and Qian Liu and Longxu Dou and Niklas Muennighoff and Zhongwei Wan and Ping Luo and Min Lin and Ngai Wong},
    booktitle={The Thirty-eighth Annual Conference on Neural Information Processing Systems},
    year={2024},
    url={https://openreview.net/forum?id=sKCKPr8cRL}
}

@inproceedings{takase2025large,
    title = "Large Vocabulary Size Improves Large Language Models",
    author = "Takase, Sho  and
      Ri, Ryokan  and
      Kiyono, Shun  and
      Kato, Takuya",
    editor = "Che, Wanxiang  and
      Nabende, Joyce  and
      Shutova, Ekaterina  and
      Pilehvar, Mohammad Taher",
    booktitle = "Findings of the Association for Computational Linguistics: ACL 2025",
    month = jul,
    year = "2025",
    address = "Vienna, Austria",
    publisher = "Association for Computational Linguistics",
    url = "https://aclanthology.org/2025.findings-acl.57/",
    doi = "10.18653/v1/2025.findings-acl.57",
    pages = "1015--1026",
    ISBN = "979-8-89176-256-5"
}

@inproceedings{merrill2024evaluating,
    title = "Evaluating $n$-Gram Novelty of Language Models Using Rusty-{DAWG}",
    author = "Merrill, William  and
      Smith, Noah A.  and
      Elazar, Yanai",
    editor = "Al-Onaizan, Yaser  and
      Bansal, Mohit  and
      Chen, Yun-Nung",
    booktitle = "Proceedings of the 2024 Conference on Empirical Methods in Natural Language Processing",
    month = nov,
    year = "2024",
    address = "Miami, Florida, USA",
    publisher = "Association for Computational Linguistics",
    url = "https://aclanthology.org/2024.emnlp-main.800/",
    doi = "10.18653/v1/2024.emnlp-main.800",
    pages = "14459--14473"
}

@inproceedings{buck2014n,
    title = "{N}-gram Counts and Language Models from the {C}ommon {C}rawl",
    author = "Buck, Christian  and
      Heafield, Kenneth  and
      van Ooyen, Bas",
    editor = "Calzolari, Nicoletta  and
      Choukri, Khalid  and
      Declerck, Thierry  and
      Loftsson, Hrafn  and
      Maegaard, Bente  and
      Mariani, Joseph  and
      Moreno, Asuncion  and
      Odijk, Jan  and
      Piperidis, Stelios",
    booktitle = "Proceedings of the Ninth International Conference on Language Resources and Evaluation ({LREC}'14)",
    month = may,
    year = "2014",
    address = "Reykjavik, Iceland",
    publisher = "European Language Resources Association (ELRA)",
    url = "https://aclanthology.org/L14-1074/",
    pages = "3579--3584"
}

@inproceedings{brants2007large,
    title = "Large Language Models in Machine Translation",
    author = "Brants, Thorsten  and
      Popat, Ashok C.  and
      Xu, Peng  and
      Och, Franz J.  and
      Dean, Jeffrey",
    editor = "Eisner, Jason",
    booktitle = "Proceedings of the 2007 Joint Conference on Empirical Methods in Natural Language Processing and Computational Natural Language Learning ({EMNLP}-{C}o{NLL})",
    month = jun,
    year = "2007",
    address = "Prague, Czech Republic",
    publisher = "Association for Computational Linguistics",
    url = "https://aclanthology.org/D07-1090/",
    pages = "858--867"
}

@inproceedings{liu2024infinigram,
    title={Infini-gram: Scaling Unbounded n-gram Language Models to a Trillion Tokens},
    author={Jiacheng Liu and Sewon Min and Luke Zettlemoyer and Yejin Choi and Hannaneh Hajishirzi},
    booktitle={First Conference on Language Modeling},
    year={2024},
    url={https://openreview.net/forum?id=u2vAyMeLMm}
}

@inproceedings{nguyen2024understanding,
    title={Understanding Transformers via N-Gram Statistics},
    author={Timothy Nguyen},
    booktitle={The Thirty-eighth Annual Conference on Neural Information Processing Systems},
    year={2024},
    url={https://openreview.net/forum?id=WCc440cUhX}
}

@misc{google2025gemma,
    title = {Gemma 3n},
    author = {{Google Team}},
    year = 2025,
    url = {https://ai.google.dev/gemma/docs/gemma-3n},
}

@inproceedings{
    tseng2026l,
    title={\$L{\textasciicircum}3\$: Large Lookup Layers},
    author={Albert Tseng and Christopher De Sa},
    booktitle={Forty-third International Conference on Machine Learning},
    year={2026},
    url={https://openreview.net/forum?id=A7LiBHEsOo}
}

@inproceedings{
    sadhukhan2026stem,
    title={{STEM}: {SCALING} {TRANSFORMERS} {WITH} {EMBEDDING} {MODULES}},
    author={Ranajoy Sadhukhan and Sheng Cao and Harry Dong and Changsheng Zhao and Attiano Purpura-Pontoniere and Yuandong Tian and Zechun Liu and Beidi Chen},
    booktitle={The Fourteenth International Conference on Learning Representations},
    year={2026},
    url={https://openreview.net/forum?id=gufRimweSQ}
}

@article{zheng2026lngram,
      title={Lngram: N-gram Conditional Memory in Latent Space}, 
      author={Yunao Zheng and Guoyang Xia and Xiaojie Wang and Lei Ren},
      year={2026},
      journal={arXiv preprint arXiv:2605.24869}
}

@inproceedings{bernstein2005poly1305,
  author    = {Daniel J. Bernstein},
  title     = {The Poly1305-{AES} Message-Authentication Code},
  booktitle = {Fast Software Encryption: 12th International Workshop, FSE 2005, Paris, France, February 21--23, 2005, Revised Selected Papers},
  editor    = {Henri Gilbert and Helena Handschuh},
  series    = {Lecture Notes in Computer Science},
  volume    = {3557},
  pages     = {32--49},
  publisher = {Springer},
  year      = {2005},
  isbn      = {3-540-26541-4}
}

@article{carter1979universal,
    title = {Universal classes of hash functions},
    journal = {Journal of Computer and System Sciences},
    volume = {18},
    number = {2},
    pages = {143-154},
    year = {1979},
    issn = {0022-0000},
    doi = {https://doi.org/10.1016/0022-0000(79)90044-8},
    url = {https://www.sciencedirect.com/science/article/pii/0022000079900448},
    author = {J.Lawrence Carter and Mark N. Wegman}
}

@article{Bhattacharyya2025poly,
title = {Polynomial hashing over prime order fields},
journal = {Advances in Mathematics of Communications},
volume = {19},
number = {1},
pages = {337-378},
year = {2025},
issn = {1930-5346},
doi = {10.3934/amc.2024001},
url = {https://www.aimsciences.org/article/id/65a63bba92d3ad47ddcdd5f0},
author = {Sreyosi Bhattacharyya and Kaushik Nath and Palash Sarkar},
}

@INPROCEEDINGS{degabriele2024sok,
  author={Degabriele, Jean Paul and Gilcher, Jan and Govinden, Jérôme and Paterson, Kenneth G.},
  booktitle={2024 IEEE Symposium on Security and Privacy (SP)}, 
  title={SoK: Efficient Design and Implementation of Polynomial Hash Functions over Prime Fields}, 
  year={2024},
  volume={},
  number={},
  pages={3128-3146},
  doi={10.1109/SP54263.2024.00132}}

@inproceedings{svenstrup2017hash,
author = {Svenstrup, Dan and Hansen, Jonas Meinertz and Winther, Ole},
title = {Hash embeddings for efficient word representations},
year = {2017},
isbn = {9781510860964},
publisher = {Curran Associates Inc.},
address = {Red Hook, NY, USA},
booktitle = {Proceedings of the 31st International Conference on Neural Information Processing Systems},
pages = {4935–4943},
numpages = {9},
location = {Long Beach, California, USA},
series = {NIPS'17}
}

@inproceedings{loshchilov2018decoupled,
    title={Decoupled Weight Decay Regularization},
    author={Ilya Loshchilov and Frank Hutter},
    booktitle={International Conference on Learning Representations},
    year={2019},
    url={https://openreview.net/forum?id=Bkg6RiCqY7},
}


\end{document}